\documentclass[preprint,12pt,authoryear]{elsarticle}

\usepackage{tabularx} 
\usepackage{booktabs}   
\usepackage{makecell}   
\usepackage{amssymb,amsfonts,amsmath,adjustbox,etoolbox,graphicx,subfigure,algorithm,algorithmicx,multirow,multicol,float,csquotes,xcolor}
\usepackage{algorithmicx}      
\usepackage[noend]{algpseudocode} 
\usepackage{xspace} 

\journal{Expert Systems with Applications}
\begin{document}
\begin{frontmatter}

\title{MSConv: Multiplicative and Subtractive Convolution for Face Recognition}

\author[label1]{Si Zhou}
\ead{sizhou@stu2023.jnu.edu.cn}

\author[label2]{Yain-Whar Si}
\ead{fstasp@um.edu.mo} 

\author[label3]{Xiaochen Yuan}
\ead{xcyuan@mpu.edu.mo}

\author[label1]{Xiaofan Li}
\ead{lixiaofan@jnu.edu.cn}  

\author[label1]{Xiaoxiang Liu}
\ead{tlxx@jnu.edu.cn} 

\author[label1]{Xinyuan Zhang}
\ead{zhangxy@jnu.edu.cn}  

\author[label1]{Cong Lin}
\ead{conglin@jnu.edu.cn} 

\author[label1]{Xueyuan Gong\corref{cor1}}  
\ead{xygong@jnu.edu.cn}  

\cortext[cor1]{Corresponding author: Xueyuan Gong}  

\affiliation[label1]{organization={School of Intelligent Systems Science and Engineering, Jinan University}, 
            state={Guangdong},
            country={China}}

\affiliation[label2]{organization={Department of Computer and Information Science, University of Macau}, 
            city={Macau},
            country={China}}

\affiliation[label3]{organization={Faculty of Applied Sciences, Macau Polytechnic University}, 
            city={Macau},
            country={China}}

\begin{abstract}
In Neural Networks, there are various methods of feature fusion. Different strategies can significantly affect the effectiveness of feature representation, consequently influencing the model’s ability to extract representative and discriminative features. In the field of face recognition, traditional feature fusion methods include feature concatenation and feature addition. Recently, various attention mechanism-based fusion strategies have emerged. However, we found that these methods primarily focus on the important features in the image, referred to as salient features in this paper, while neglecting another equally important set of features for image recognition tasks, which we term differential features. This may cause the model to overlook critical local differences when dealing with complex facial samples. Therefore, in this paper, we propose an efficient convolution module called MSConv (Multiplicative and Subtractive Convolution), designed to balance the model’s learning of salient and differential features. Specifically, we employ multi-scale mixed convolution to capture both local and broader contextual information from face images, and then utilize Multiplication Operation (MO) and Subtraction Operation (SO) to extract salient and differential features, respectively. Experimental results demonstrate that by integrating both salient and differential features, MSConv outperforms models that only focus on salient features. 
\end{abstract}


\end{frontmatter}

\section{Introduction}
\label{sc:intro}
In recent years, convolutional neural networks (CNNs) have obtained widespread applications in computer vision tasks \citep{Russakovsky2015} due to its ability in obtaining representative features. Among these, effective feature fusion methods can maximize the utilization of initially extracted features to improve the recognition accuracy of models, posing significant challenges in developing more reasonable and efficient feature fusion strategies. To address these challenges, various feature fusion strategies have been explored to enhance feature extraction outcomes. Traditional methods such as feature addition and feature concatenation can improve the performance of models to some extent. 

\begin{figure}[!t]
    \centering
    \includegraphics[width=0.8\columnwidth]{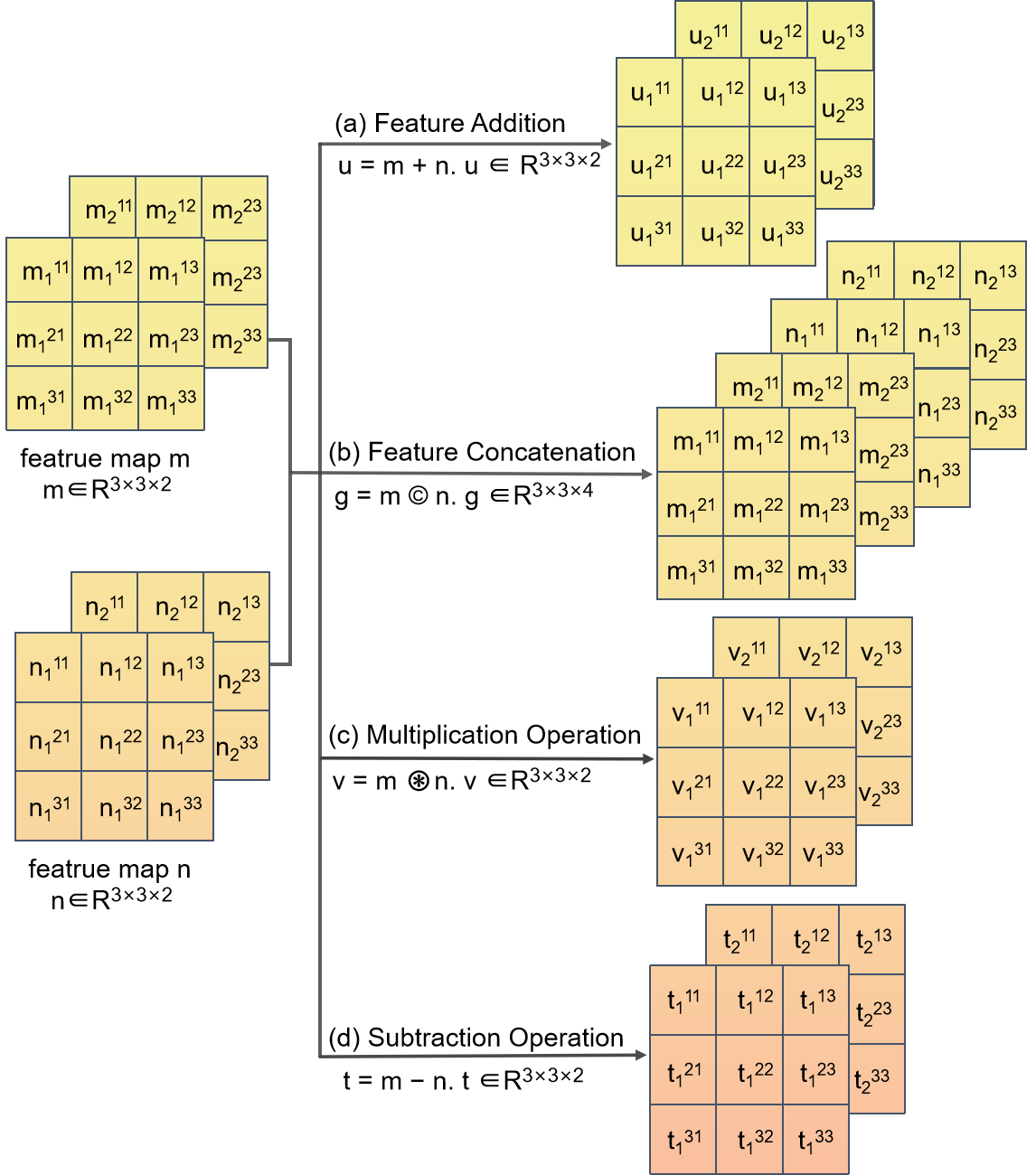} 
    \caption{Four feature fusion methods.  The symbol \enquote{+} denotes element-wise addition, \enquote{$\circledast$} denotes element-wise multiplication, \enquote{--} denotes element-wise subtraction, and \enquote{\textcircled{c}} denotes channel-wise concatenation of feature maps. Here, $ u $, $ g $, $ v $, and $ t $ represent the fused feature maps. Note that only the feature concatenation operation increases the channel dimension.}
    \label{fg:11}
\end{figure}

Their operations are straightforward. Feature addition generates a new feature tensor by adding two tensors element-wise, as shown in the Fig. \ref{fg:11}(a). Feature addition increases the amount of information in the features. Specifically in residual connections, feature addition promotes gradient descent, enhances training stability, and alleviates gradient vanishing problems in deep networks. Feature addition requires no extra trainable parameters and entails extremely low computational overhead. Within the same inference time, the network can quickly integrate multiple features, making multi-layer and cross-layer fusion feasible even in deeper architectures, thereby consolidating information at multiple scales. Although addition is a linear operator, element-wise summation fuses multiple feature maps into a tensor of the same dimensions \citep{He2016}. If the same location across different feature maps exhibits high activation, the sum results in an even stronger signal at that position, highlighting the importance of features in that region. Conversely, if only a few feature maps have high responses while others are low, the summation provides a smoothing effect that partially filters out random noise.

Feature concatenation combines two or more tensors along a specific dimension to form a larger tensor, as shown in Fig. \ref{fg:11}(b), increasing the number of channels or feature dimensions in deep learning models to better capture the relationships between different features in subsequent layers. Feature concatenation merges multiple feature streams along the channel or vector dimension, thereby more fully preserving potential complementary information across different channels, network branches, or hierarchical levels \citep{huang2017}. For face recognition tasks, which require the capture of various subtle differences, this approach allows the model to combine local and global features more flexibly in subsequent stages \citep{schroff2015facenet}. Moreover, because the concatenated feature space becomes larger, the model can represent inputs from a richer-dimensional perspective, helping to maintain stable discrimination ability under variations in lighting, pose, and expression.

Traditional feature fusion techniques often focus on high-response features, but in tasks like face recognition, where fine-grained discrimination is essential, subtle yet decisive details can be crucial for distinguishing highly similar faces. Overreliance on salient features tends to obscure these details, limiting recognition accuracy in complex scenarios. Additionally, traditional methods are typically linear, and they do not offer a dynamic mechanism to adjust attention across different feature channels or spatial regions based on the input distribution. Even with fluctuating salience, the model treats all features equally, making it difficult to distinguish important signals from noise. In deep neural networks, features often exhibit nonlinear relationships, especially with multi-level representations or multimodal inputs. Simple fusion approaches fail to exploit higher-order correlations among feature streams, resulting in suboptimal performance in challenging settings.

To address these challenges, attention mechanisms have been introduced during the fusion phase. Unlike traditional feature fusion methods, attention mechanisms learn channel or position weights during training that adapt to the task and data distribution. In contrast, attention mechanisms assign lower weights to noisy features, allowing the model to focus on the most useful ones during inference. When faced with diverse inputs, such as lighting, expression, angle, or occlusion variations, the model can automatically adjust which aspects to emphasize.

On the channel dimension, the network aggregates information globally, average or max pooling, then generates channel attention weights through trainable layers. These weights are multiplied with the original features during the forward pass, amplifying important channels while suppressing less relevant ones \citep{Hu2018squeeze, Wang2020}. On the spatial dimension, attention assigns dynamic weights to each pixel or local region of the feature map. By pooling channel information into a 2D representation, followed by lightweight convolution or MLP processing, the model infers a spatial weight matrix \citep{Zhuang2024}. This weight is multiplied with the feature map, highlighting key regions such as the eyes or mouth, while suppressing irrelevant background. However, attention mechanisms exhibit inherent limitations, such as over-compression of features during attention weight computation and neglect of spatial positional relationships \citep{Hu2018squeeze}. Moreover, sophisticated attention mechanisms like CoTAttention and Multi-Head Self-Attention often impose high computational complexity, demanding substantial computational resources \citep{vaswani2017attention, hou2021coordinate}.

Additionally, all these prior mentioned methods share a common drawback, they focus heavily on salient features while neglecting the importance of differential features. Traditional feature fusion techniques often focus on the most intuitive, high-response features. However, in face recognition tasks, where fine-grained discrimination is crucial, some subtle but decisive details can be critical for distinguishing highly similar faces. Overreliance on salient features tends to bury these fine details, limiting the model’s recognition accuracy in high-similarity or complex scenarios. Moreover, traditional fusion methods are typically linear or nearly linear, offering no mechanism to dynamically adjust attention across different feature channels or spatial regions based on the input distribution. Even if the salience of certain dimensions or positions fluctuates significantly, the model still treats all features equally, making it difficult to effectively distinguish important signals from noise or redundancy.In deep neural networks, features often exhibit nonlinear and complex relationships, especially when involving multiple levels of representation or multimodal inputs. Simple fusion approaches can only achieve shallow aggregation and fail to fully exploit higher-order correlations among multiple feature streams, resulting in suboptimal performance in challenging face recognition settings.

\begin{figure}[!t]
	\centering
	\includegraphics[width=\columnwidth]{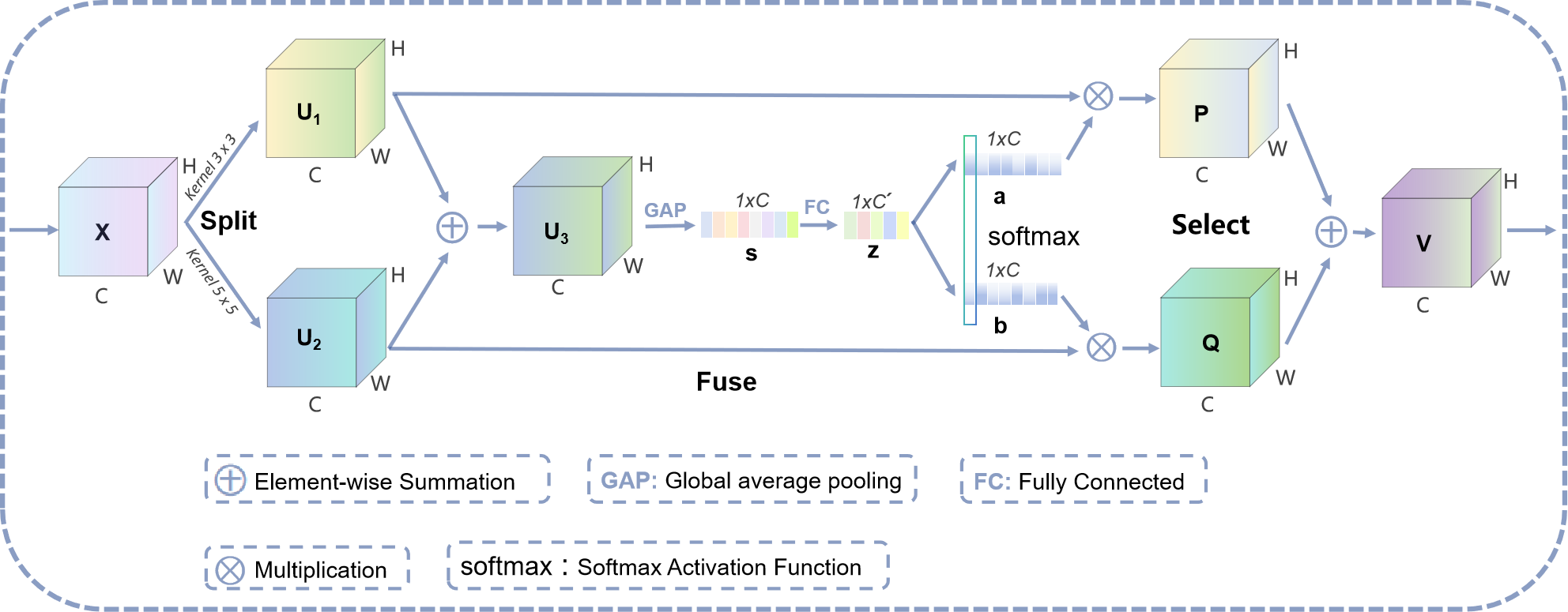}
	\caption{Selective Kernel Convolution.}
	\label{fg:sknet}
\end{figure}

Building on these studies, we integrate the advantages of attention mechanisms and mixed operators (MO and SO) into MSConv, thereby preserving the efficient utilization of salient features while also accounting for differential features. This approach yields more comprehensive and accurate recognition performance overall. In this paper, inspired by the works of Rewrite the Stars(StarNet) \citep{Ma2024} and Selective Kernel Networks(SKNet) \citep{Li2019}, we design a novel feature fusion strategy aiming to balance the learning of salient features and differential features to enhance feature expressiveness while diversifying the development of feature fusion strategies. In the SKNet model, the authors firstly use the multi-scale convolution method for initial feature extraction, then employ feature addition to fuse these features to obtain a new feature output. Finally, they use the attention mechanism to further fuse the features, as shown in Fig. \ref{fg:sknet}. 

We found that the softmax activation function in the SKNet model can be further transformed into the sigmoid activation function, with more detailed steps described in Section \ref{sc:pa}. After reformulating the equations, the model structure undergoes significant changes, and within the transformed structure, the element-wise subtraction operation is discovered and we name it SO, as shown in the Fig. \ref{fg:11}(d). We introduces the SO to help the model extract differential features. Traditional feature fusion methods typically focus on salient features, while differential features can capture subtle variations that may be overlooked by conventional methods. Differential features refer to those that capture the unique differences between feature maps, usually reflecting the distinctiveness or specific state of the samples, which is crucial in classification tasks where the samples are visually similar but belong to different categories. Particularly in face recognition tasks, differential features capture the differences between different face samples or the variations in the same object under different lighting and angle conditions. Differences between individuals, such as the shape of the nose and the distance between the eyes, are critical for recognition \citep{Zangeneh2020}. By learning these differential features, the model can better handle noise, occlusion, or background changes, allowing it to focus on the most important information. In summary, differential features provide additional discriminative information, and by integrating both differential features and salient features, the model can achieve a more comprehensive and effective understanding of face samples, leading to more stable and reliable performance in practical applications.

Meanwhile, in the StartNet model, the authors provide a deep interpretation of the \enquote{star operation} (element-wise multiplication). Inspired by this, we propose a novel element-wise multiplication, which differs from the SKNet model's approach of performing fully connected operations on feature maps separately before using element-wise multiplication as shown in the Fig. \ref{fg:11}(c). Instead, we multiply two different features obtained through multi-scale convolution methods, naming this operation MO, and then introduce it into the transformed SKNet model structure, which is the MSConv proposed in this paper. Our MSConv can be embedded into various architectures without the need for additional modifications, aiming to efficiently utilize features and enhance their expressiveness. Experiments have proven the effectiveness of our work. In summary, the main contributions in this paper can be concluded as follows:

\begin{itemize}
	\item To the best of our knowledge, we are the first to formally propose the importance of learning differential features and introduce the SO to extract differential features. SO suppresses similar or identical parts of two feature maps, allowing the model to focus more on truly distinctive information, which is beneficial for removing redundant information and reducing noise.
	\item We propose a novel method, MO, for learning salient features. When the feature values in the same region across different feature maps are large, MO further amplifies these important features. When the feature values are small, MO reduces their impact, helping the model ignore these less important features. When there is a large difference in feature values, MO stabilizes them by either highlighting the salient features or diminishing the unclear ones.
	\item We design a plug-and-play operation named MSConv combining MO and SO to replace standard convolution for operating on a variety of backbone CNNs. It turns out that MSConv can enhance the model performance on challenging tasks.
\end{itemize}

The remainder of this paper is organized as follows: The related work is reviewed in Section \ref{sc:rw}. Next, MSConv, SO and MO are introduced in Section \ref{sc:pa}. The experimental results are discussed in Section \ref{sc:exp}. Finally, Section \ref{sc:conc} concludes the paper.

\section{Related Work}
\label{sc:rw}
Feature fusion is an important research direction for neural networks in computer vision tasks, particularly in face recognition tasks. This section reviews existing feature fusion methods, categorizing them into traditional methods and attention mechanisms, and provides further analysis of these methods.

Early feature fusion methods mainly focused on feature addition and feature concatenation. Feature addition involves element-wise addition of two feature maps to generate a feature map, facilitating residual connections. These connections promote gradient descent, enhance training stability and alleviate the vanishing gradient problem, as demonstrated by \citep{He2016} in Residual Network(ResNet) and \citep{huang2017densely} in Densely Connected Convolutional Network(DenseNet). Feature concatenation, on the other hand, merges feature maps from different sources or levels along a specific dimension, increasing the number or dimensionality of features to capture relationships between different features. For example, Long et al. \citep{Long2015} used feature concatenation in fully convolutional networks (FCNs) for semantic segmentation, significantly improving performance. Although this method can integrate information from multiple sources, it often leads to a significant increase in model complexity. However, despite the effectiveness of addition and concatenation, these traditional methods primarily focus on enhancing salient features, often neglecting the importance of differential features. This limitation has driven researchers to explore more advanced feature fusion strategies.

To avoid applying the same weight to all features, researchers have introduced attention mechanisms during the fusion phase, emphasizing fine-grained differences and ensuring the model maintains enough discriminative power even under highly similar faces. Unlike averaging-style traditional fusion, attention mechanisms leverage learnable weights during training to highlight crucial channels or spatial positions and suppress noisy features. Such advanced fusion strategies enable the model to focus on genuinely discriminative regions or dimensions. 

\begin{figure}[!t]
	\centering
	\includegraphics[width=\columnwidth]{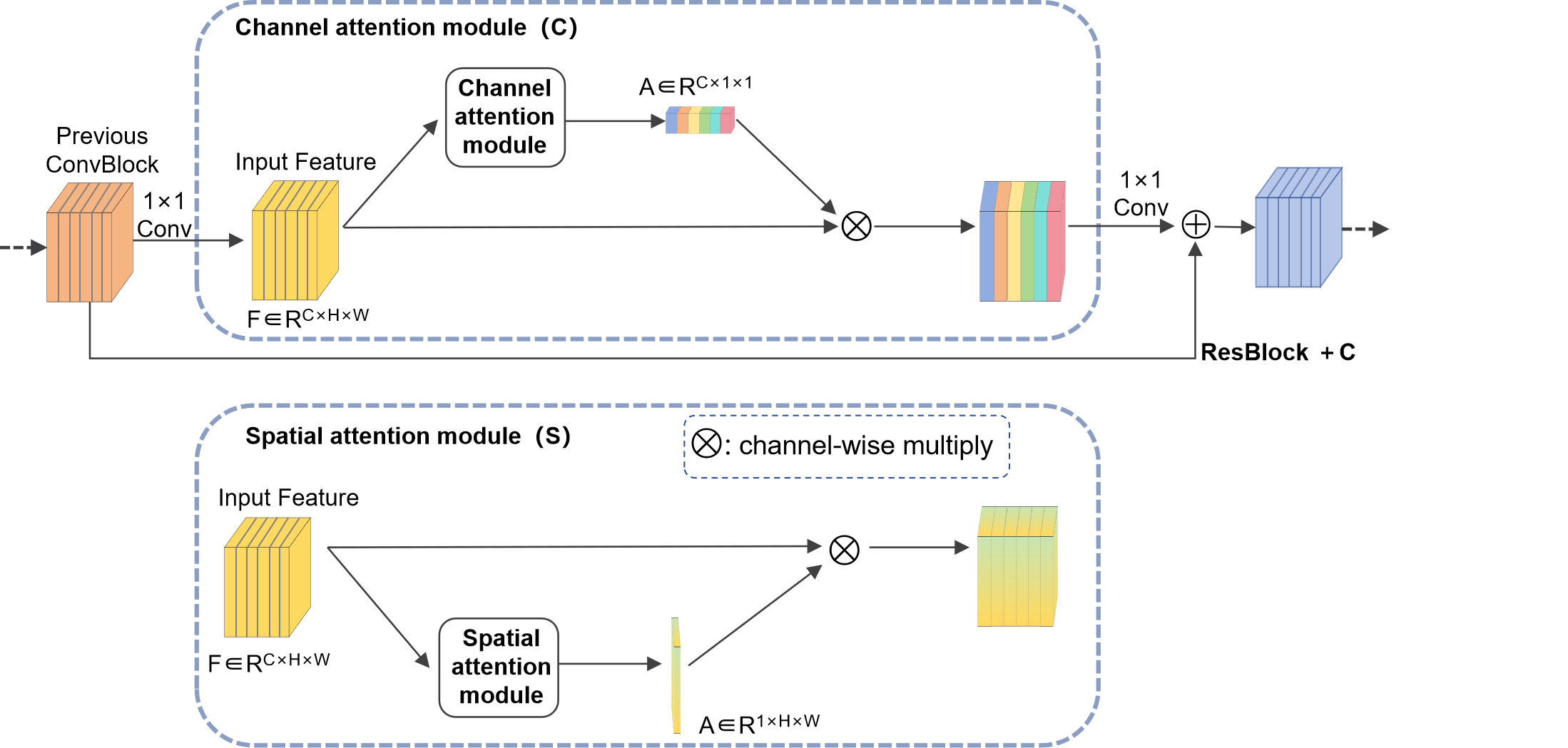}
	\caption{Two different attention mechanisms and the insertion points in foundational models (ResNet). \textbf{C} denotes the Channel Attention Module, and \textbf{S} denotes the Spatial Attention Module.}
	\label{fg:figure3}
\end{figure}

As a complex feature fusion strategy, Common attention mechanisms include channel attention, spatial attention and branch attention. Recent works have shown the combined use of these attention mechanisms, such as the commonly used combination of channel attention and spatial attention mechanisms. Attention modules are typically embedded into key hierarchical structures of CNNs. A common insertion point is  after the convolutio layers. For instance, in the ResNet architecture, commonly used channel attention modules are often inserted between the two 1×1 convolutional layers (equivalent to fully connected layers) within a residual block, as shown in Fig. \ref{fg:figure3}.

\begin{enumerate}[(1)]
\item Channel attention mechanisms assign weights to feature map channels, helping the model focus on critical ones. \citep{Hu2018squeeze} proposed Squeeze-and-Excitation Network(SENet), which uses \enquote{squeeze} and \enquote{excitation} modules to adaptively adjust channel importance, enhancing image classification by capturing and reweighting global information. \citep{Wang2020} introduced Efficient Channel Attention Network(ECA-Net), an efficient mechanism that avoids dimensionality reduction, minimizing information loss. ECA-Net improves the balance between complexity and performance with a local interaction mechanism that provides effective attention at minimal computational cost.
\item Spatial attention mechanisms assign weights to different regions of the feature map, helping the model focus on the most relevant image areas. \citep{Wang2017} introduced spatial attention in their ResNet, embedding it into residual blocks to improve image classification by focusing on critical regions at various levels. Other models also utilize spatial attention, such as Recurrent Attention Model(RAM) \citep{Xu2015}, which uses a recurrent approach to focus on regions during image captioning. Spatial transformer networks(STN) \citep{Jaderberg2015}, which applies a learnable transformation for enhanced feature extraction; and Non-local Neural Networks(NLNN) \citep{Wang2018}, which capture long-range dependencies for better global information.

\item Branch attention mechanisms enhance model flexibility and accuracy by combining feature fusion methods across branches. Conditionally Parameterized Convolutions(CondConv) \citep{Yang2019} dynamically generates convolutional kernel weights based on input, adapting feature extraction without increasing computational complexity. Dynamic Convolution \citep{Chen2020} selects and combines kernels dynamically, reducing complexity while maintaining strong performance across diverse data patterns. Both methods improve model adaptability and robustness in complex scenarios by leveraging multi-branch structures.

\item Combining channel and spatial attention enhances feature map representation. \citep{Woo2018} and \citep{Zhuang2024} respectively introduced Convolutional Block Attention Module(CBAM) and Frequency Regulated Channel-Spatial Attention Module(FReCSA), both of which utilize channel and spatial attention to improve key feature representation and overall model performance. \citep{Li2023} proposed Spatial and Channel Convolution(SCConv), which uses spatial and channel reconstruction strategies to refine feature layout and relationships, boosting the model's ability to understand complex visual patterns, especially in high-resolution images.
\end{enumerate}

Despite these advanced methods pushing the boundaries of feature fusion, they still have some limitations. Many methods, although effective, involve higher computational complexity and longer training times. Furthermore, most existing methods still primarily focus on salient features, often overlooking differential features. This imbalance can lead to incomplete feature representation, limiting the model's ability to capture all relevant information. Differential features capture subtle differences in visually similar but categorically distinct face samples, whereas traditional feature fusion methods primarily focus on salient features, often missing these variations. Integrating differential and salient features provides the model with a more comprehensive feature representation, enabling a more accurate understanding of face samples. 

To address these limitations, we propose MSConv. MSConv is designed to balance the model's attention to salient features and differential features, enhancing the comprehensiveness and accuracy of feature learning. By combining MO and SO, MSConv ensures more complete and expressive feature representation. This work not only enhances the understanding of feature fusion strategies but also aids future research in further refining these techniques.

\section{Proposed Approaches}
\label{sc:pa}
\subsection{Salient Features \textit{vs.} Differential Features}
Salient features help the model extract key characteristics from most faces, enabling basic identity recognition under typical lighting and angle variations. However, when the model encounters highly similar samples, relying solely on these features becomes insufficient, leading to confusion. In such cases, the advantages of salient features become a bottleneck when distinguishing between faces with extremely high similarity. In contrast, differential features (subtle facial textures) play a decisive role in distinguishing highly similar faces but are highly sensitive to changes in the external environment and individual conditions. If the model relies too heavily on these differential features, they can become unstable in scenarios such as low lighting, backlighting, or drastic facial expression changes. Therefore, practical face recognition tasks require the model to simultaneously capture and integrate both salient and differential features.

\subsection{MSConv Module}
\label{ssc:itt}
\begin{figure}[!t]
	\centering
	\includegraphics[width=\columnwidth]{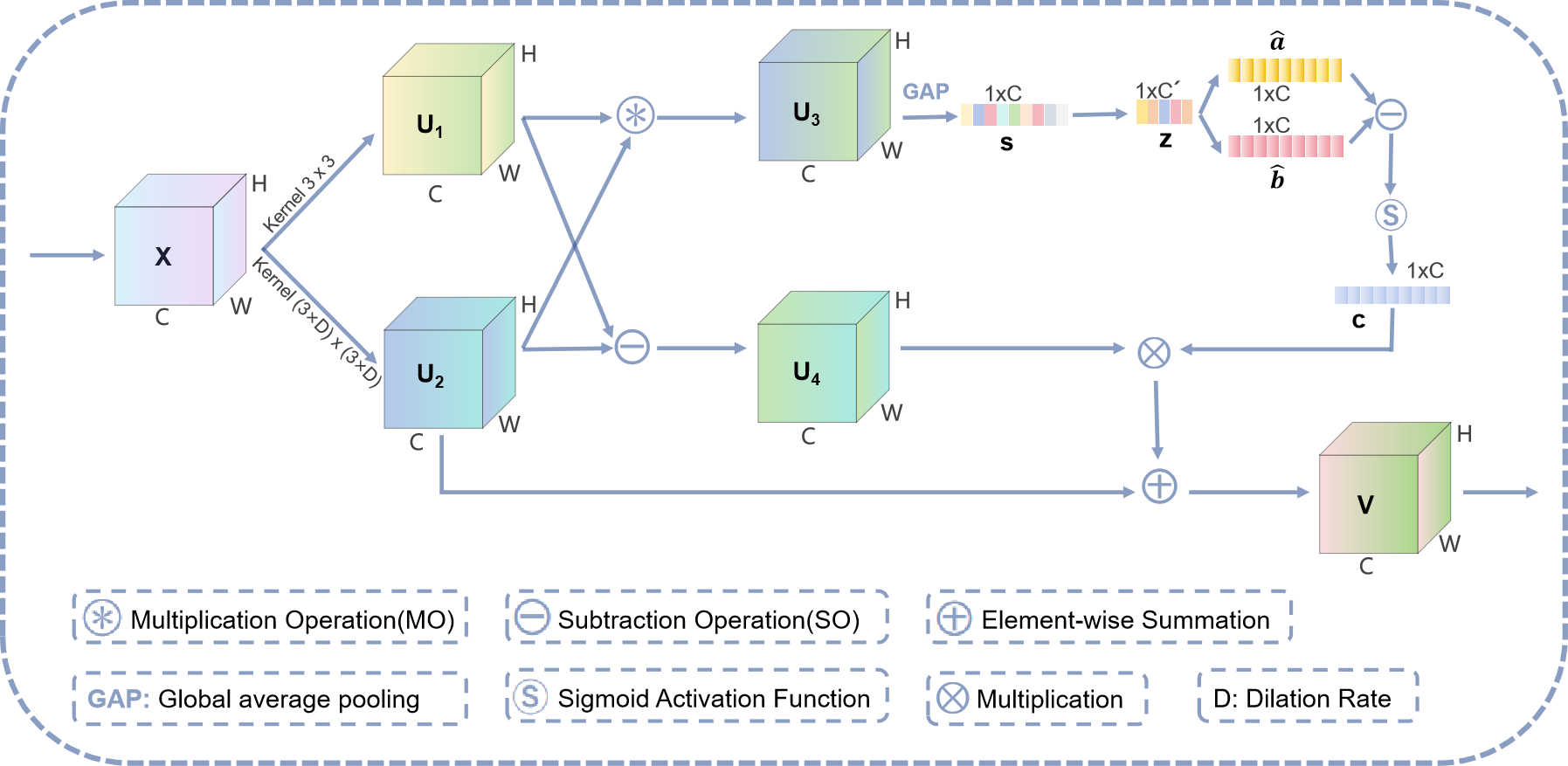}
	\caption{The architecture of MSConv.}
	\label{fg:msnet}
\end{figure}

In order to balance the model's attention to both salient and differential features, we propose MSConv, as illustrated in Fig. \ref{fg:msnet}. Specifically, for the output features $U_{1}$ and $U_{2} \in \mathbb{R}^{H \times W \times C}$ from the multi-scale convolution, we first obtain the salient features $U_{3} \in \mathbb{R}^{H \times W \times C}$ through the MO, where $N$ is the batch size, $C$ and $C^{\prime}$ is the channel dimension, while $H$ and $W$ are the height and width of the spatial dimensions, respectively. Subsequently, we use the attention mechanism to obtain the attention weights $ \boldsymbol{c} $, and then use the SO to output the differential features $U_{4} \in \mathbb{R}^{H \times W \times C}$. More precisely, in this attention mechanism, we embed global information by applying global average pooling to generate channel-wise statistics $s \in \mathbb{R}^{1 \times C}$. A compact feature $z \in \mathbb{R}^{1 \times d}$ is then produced through a fully connected (FC) layer with dimensionality reduction to improve computational efficiency. The dimension $C^{\prime}$ is defined as Eq. \eqref{eq:c}: 
\begin{equation}
\label{eq:c}
    C^{\prime} = \max\left( \frac{C}{r}, L \right)
\end{equation}
where $r$ is a reduction ratio and $L$ denotes the minimum value of $C^{\prime}$, initialized to 32. Related vectors  $\mathbf{a}$ and $\mathbf{b}$, and the detailed operations are given in Section 3.1. Additionally, the attention weights $c \in \mathbb{R}^{1 \times C} $ are fused with the differential features $U_{4}$ to produce the output features, which are finally combined with the features $U_{2}$ to yield the final output $V \in \mathbb{R}^{H \times W \times C}$. In the MSConv module, we exploit both salient features and differential features to balance the model's learning of these equally important features and enhance the feature representation capability of CNNs. Note that the same characters in both Fig. \ref{fg:sknet} and Fig. \ref{fg:msnet} represent the same meaning.

Given that MO demonstrates the ability to highlight significant features in feature fusion, we replaced the first element-wise summation with MO in the SKConv module. For the model output $V$, the relationship derived in Eq. \eqref{eq:sce} holds:

\begin{equation}
\label{eq:sce}
    \quad
        \begin{cases} 
            \mathbf{V}^i = \mathbf{P}^i + \mathbf{Q}^i \\
            \mathbf{P}^i = \mathbf{a}^i \times \mathbf{U}_1^i \\
            \mathbf{Q}^i = \mathbf{b}^i \times \mathbf{U}_2^i \\
            \mathbf{a}^i = \frac{e^{\hat{\mathbf{a}}^i}}{e^{\hat{\mathbf{a}}^i} + e^{\hat{\mathbf{b}}^i}} \\
            \mathbf{b}^i = \frac{e^{\hat{\mathbf{b}}^i}}{e^{\hat{\mathbf{a}}^i} + e^{\hat{\mathbf{b}}^i}} \\
            \mathbf{a}^i + \mathbf{b}^i = 1
        \end{cases}
\end{equation}
where $i \in \{1, 2, ..., C\}$. Note that $\mathbf{V}^i \in \mathbb{R}^{H \times W}$ is the i-th channel of $V$ and $a^i \in \mathbb{R}$ is the i-th element of $a \in \mathbb{R}^{1 \times C}$, likewise for $P^i$, $Q^i$, $U_1^i$, $U_2^i$, $U_3^i$, $U_4^i$ and $b^i$, $c^i$, $\hat{a}^i$, $\hat{b}^i$. Note that the batch size dimension $N$ is not considered here. Additionally, $\hat{a}$ and $\hat{b}$ represents the different results of upsampling $z \in \mathbb{R}^{1 \times C'}$ to rescale the number of channels to $C$. By further deriving from Eq. \eqref{eq:sce}, we obtain Eq. \eqref{eq:scels}:

\begin{equation}
\label{eq:scels}
    \begin{aligned}
        \mathbf{V}^i &= \mathbf{a}^i \times \mathbf{U}_1^i + \mathbf{b}^i \times \mathbf{U}_2^i \\
        &= \mathbf{a}^i \times \mathbf{U}_1^i + (1 - \mathbf{a}^i) \mathbf{U}_2^i \\
        &= \mathbf{a}^i (\mathbf{U}_1^i - \mathbf{U}_2^i) + \mathbf{U}_2^i \\
        &= (\mathbf{U}_1^i - \mathbf{U}_2^i) \times \frac{e^{\hat{\mathbf{a}}^i}}{e^{\hat{\mathbf{a}}^i} + e^{\hat{\mathbf{b}}^i}} + \mathbf{U}_2^i \\
        &= \frac{\mathbf{U}_1^i - \mathbf{U}_2^i}{1 + e^{-(\hat{\mathbf{a}}^i - \hat{\mathbf{b}}^i)}} + \mathbf{U}_2^i \\
    \end{aligned}
\end{equation}
Based on the condition information provided by Eq. \eqref{eq:scels}, Eq. \eqref{eq:v} can be further derived to Eq. \eqref{eq:vc}, ultimately resulting in Eq. \eqref{eq:vcc}:
\begin{equation}
\label{eq:v}
    \quad
        \begin{cases}
        \mathbf{U}_4^i = \mathbf{U}_1^i - \mathbf{U}_2^i \\
        \mathbf{c}^i = \mathbf{a}^i - \mathbf{b}^i \\
        \textit{Sigmoid}(\mathbf{c}^i) = \frac{1}{1 + e^{-\mathbf{c}^i}}  
        \end{cases}
\end{equation}

\begin{equation}
\label{eq:vc}
    \mathbf{V}^i = \mathbf{U}_4^i \times \textit{Sigmoid}(\mathbf{c}^i) + \mathbf{U}_2^i
\end{equation}

\begin{equation}
\label{eq:vcc}
    \mathbf{V} = \mathbf{U}_4 \times \textit{Sigmoid}(\mathbf{c}) + \mathbf{U}_2
\end{equation}
Through a series of derivations, the parameter relationship represented by Eq. \eqref{eq:vcc} is the base relationship of MSConv. By replacing the element-wise summation in SKConv with MO and applying the transformation from Eq. \eqref{eq:scels}, we built the MSConv module, as shown in Fig. \ref{fg:msnet}. The corresponding pseudocode is shown in Algorithm ~\ref{alg:msconv}. This module uses a dual-branch structure, where one branch applies a traditional 3$\times$3 convolution and the other uses a dilated 3$\times$3 convolution, producing feature maps $U_1$ and $U_2$. The traditional 3$\times$3 convolution is known for its ability to capture fine local features by sliding a kernel over the image. 

\begin{algorithm}[!t]
    \caption{Algorithm of MSConv}
    \label{alg:msconv}
    \begin{algorithmic}[1]
        \Require Given a feature map $X \in \mathbb{R}^{H \times W \times C}$
        \Ensure The output feature map $V \in \mathbb{R}^{H \times W \times C}$
        \State Generate multi-scale features
        \State \quad $U_1 \in \mathbb{R}^{H \times W \times C} \leftarrow \text{Conv}_{3\times3}(X) $ 
        \State \quad $U_2 \in \mathbb{R}^{H \times W \times C} \leftarrow \text{Conv}_{5\times5}(X)$ \Comment{$U_2$, $U_2 \in \mathbb{R}^{H \times W \times C}$}
        \State Compute fusion maps:
        \State \quad $U_3 \in \mathbb{R}^{H \times W \times C} \leftarrow U_1 \times U_2$ \Comment{Element-wise multiplication}
        \State \quad $U_4 \in \mathbb{R}^{H \times W \times C} \leftarrow U_1 - U_2$ \Comment{Element-wise subtraction} 
        \State Global Average Pooling:
        \State \quad $s \leftarrow \frac{1}{H \times W} \sum_{i=1}^{H}\sum_{j=1}^{W} U_3(i,j,:)$ \Comment{$s \in \mathbb{R}^{1 \times C}$}
        \State Dimension reduction:
        \State \quad $z \leftarrow \text{Conv}_{1\times1}(s)$ \Comment{$z \in \mathbb{R}^{1 \times d}$, $d = max(C/r, L=32)$ and r is a reduction ratio}
        \State Dimension expansion:
        \State \quad $[\hat{a}, \hat{b}] \leftarrow \text{Conv}_{1\times1}(z)$ \Comment{$\hat{a}, \hat{b} \in \mathbb{R}^{1 \times d}$}
        \State Compute the attention weight vector:
        \State \quad $c \leftarrow \sigma(\hat{\alpha} - \hat{b})$ \Comment{$\sigma$: Sigmoid function}
        \State Feature fusion:
        \State \quad $V \leftarrow U_2 + c \times U_4$  \Comment{$V \in \mathbb{R}^{H \times W \times C}$}
        \State \Return{$V$}
    \end{algorithmic}
\end{algorithm} 

However, its limited receptive field makes it less effective in capturing global information. Dilated convolution is no longer merely a tool for expanding the receptive field to compensate for the limitations of traditional convolution. With the development of deep learning, dilated convolution has evolved into a more strategic tool by incorporating multi-scale aggregation, sparse connections and integration with attention mechanisms to maximize feature capture while reducing computational costs. Specifically, dilated convolution achieves multi-scale feature aggregation by applying different dilation rates across various layers, allowing the model to focus on both local details and broader contextual information simultaneously. It also leverages sparse connections by inserting gaps between convolutional kernel elements to expand the receptive field, thereby reducing computational complexity while maintaining the ability to capture information over a wide range\citep{Yu2015}. Additionally, the combination of dilated convolution with attention mechanisms further enhances the model's focus on important regions, enabling the model to capture both more global contextual information and key local details. Despite the expanded receptive field, the use of sparse connections ensures that dilated convolution maintains efficient feature extraction without significantly increasing computational costs, making it highly efficient and capable of powerful feature extraction in deep learning. Together, they provide the model with a powerful multi-scale feature extraction capability, improving its ability to handle complex scenes by capturing both fine details and overall structure.

\subsection{Multiplication Operation(MO)}
\label{ssc:it}
Element-wise multiplication exhibits a bilinear property, introducing non-linearity into feature interactions compared to the single-linearity of feature addition. This allows for mutual interactions between different features. Without introducing additional parameters, element-wise multiplication emphasizes collaborative or competitive relationships between features at the same spatial location by treating the two feature maps as mutual weights, thereby revealing deeper feature interactions. Specifically, during forward propagation, the output at each position is jointly determined by both feature maps, rather than relying on manually set parameters or those derived from additional complex computations. When the other feature map also has a large value at the same location, the multiplication operation further amplifies these strong responses. In other words, the output is conditional—significant enhancement occurs only when both feature maps are highly activated at the same spatial position. If the feature value at a corresponding position in either map is small, the multiplication result remains minimal, preventing uncertain information from being excessively amplified in the final features.

Furthermore, during the backpropagation process, the multiplication operation dynamically adjusts feature weights based on the inputs. As shown in Equation \eqref{eq:forward}, when feature $\mathbf{y}$ is large, the gradient with respect to feature $\mathbf{x}$ is correspondingly amplified, and vice versa. This implies that when a feature value is in a high-response state, it further amplifies the gradient of the other feature, leading to better optimization during training and faster learning of salient features.

\begin{equation}
\label{eq:forward}
    \left\{
    \begin{aligned}
        &\text{Out} = x \cdot y, \\
        &\frac{\partial \text{Out}}{\partial x} = y, \\
        &\frac{\partial \text{Out}}{\partial y} = x.
    \end{aligned}
    \right.
\end{equation}

\subsection{Subtraction Operation(SO)}
We propose incorporating MO into the MSConv module to help the model extract differential features during training. Through element-wise subtraction, SO eliminates out similar information across different feature maps, reducing redundancy and directing the model's attention to genuinely distinct elements. 

\begin{figure}[!t]
	\centering
	\includegraphics[width=0.65\textwidth, height=0.45\textheight, keepaspectratio]{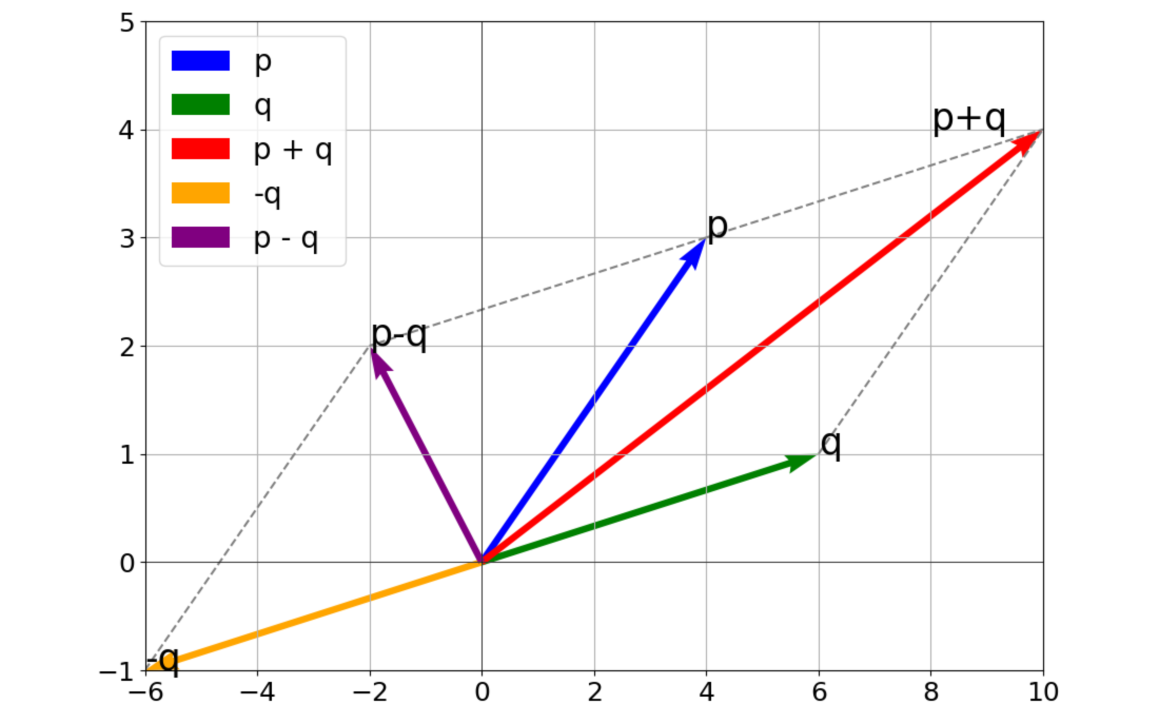}
	\caption{Vector Addition Operations and Subtraction Operations. The dashed lines in the figure are auxiliary lines.}
	\label{fg:vs}
\end{figure}

SO suppresses redundant shared components between different features and emphasizes discriminative features. As shown in Fig. \ref{fg:vs}, we analyze the addition and subtraction of vectors $p$ and $q$ in a 2D space to illustrate feature relationships in high dimensional spaces. The vector $p + q$ aligns closely with the directions of the original vectors $p$ or $q$, whereas $p - q$ exhibits a significantly different direction. Additionally, the magnitudes of $p + q$ and $p - q$ differ markedly, indicating that element-wise addition amplifies shared information between features. During training, redundant similarities accumulate over time, scattering the attention of model and weakening its focus on critical features. In contrast, the element-wise subtraction $p - q$ eliminates shared components between feature maps, directing the attention to more discriminative patterns and reducing interference from redundancy.

Reducing Noise. In most practical scenarios, noise is independent and random \citep{Haykin2002}. Consider two feature vectors $\mathbf{\xi}_1$ and $\mathbf{\xi}_2$ containing noise components $\mathbf{N}_1$ and $\mathbf{N}_2$, respectively. Following the noise modeling assumptions in \citep{Bishop2006, Zhang2018}, we assume the noise follows a normal distribution: $\mathbf{N}_1, \mathbf{N}_2 \sim \mathcal{N}(\mu, \sigma^2)$. If the noise sources are similar, the distribution characteristics of $\mathbf{N}_1$ and $\mathbf{N}_2$ are closely aligned. The compositions of the two feature vectors are expressed as shown in Eq. \eqref{eq:scess}:  
\begin{equation}
\label{eq:scess}
    \begin{aligned} 
    \mathbf{\xi}_1 = \mathbf{S}_1 + \mathbf{N}_1 \\
    \mathbf{\xi}_2 = \mathbf{S}_2 + \mathbf{N}_2
    \end{aligned} 
\end{equation} 
\begin{equation}
\label{eq:sn}
    \begin{aligned}
    \quad\quad\quad\quad\mathbf{Z} &= \mathbf{\xi}_1 - \mathbf{\xi}_2 \\
    &= (\mathbf{S}_1 - \mathbf{S}_2) - (\mathbf{N}_1 - \mathbf{N}_2) \\
    &= \mathbf{S} - \mathbf{N}
    \end{aligned} 
\end{equation} 
where $S_1$, $S_2$ and $S$ represent the true components, and $N_1$, $N_2$ as well as $N$ represent the noise components. Applying a subtraction operation yields Eq.\eqref{eq:sn}. Here, $S$ corresponds to the signal difference, while $N$ represents the noise difference. Since ${N}_1$ and ${N}_2$ are independent and identically distributed, the noise difference $N$ has a mean of zero and a variance of $2\sigma^2$. After subtraction, the signal difference $S$ is preserved, whereas the residual noise $N$ exhibits an expected value of zero, achieving effective separation of information from noise.

SO reduces redundant similar features in different features and suppresses the influence of noise, allowing the model to focus more on the more discriminative features in face samples. This helps reduce the overlap between different classes in the feature space, making the feature distribution of each class clearer, thus enabling the model to more easily focus on the different features between the feature maps during the learning process.

\section{Experiments}
\label{sc:exp}
\subsection{Experimental settings}
\label{ssc:hsa}
All experiments in this paper are implemented using Pytorch, and mixed-precision \citep{Micikevicius2017} is employed to save GPU memory and accelerate training. We follow \citep{Deng2019, Wang2018non} to set the hyper-parameters of margin-based softmax cross-entropy loss(hereinafter referred to as softmax loss) and adopt flip data augmentation. We use customized ResNet \citep{He2016deep} as the backbone. For the training of CNN models, the default batch size is set as 128. In our experiments, we employed the Stochastic Gradient Descent (SGD) optimizer to train the model. The initial learning rate was set to 0.02, and we utilized a momentum coefficient of 0.9 to accelerate convergence and reduce oscillations during training. Additionally, to prevent overfitting, a weight decay regularization term with a coefficient of 5e-4 was applied. The size of the images used for training and evaluation 112 $\times$ 112 $\times$ 3, and the dimension of the face features is 512. All pixel values of the images used for training and evaluation are normalized to between -1 and 1.

Our experiments are conducted on a computer equipped with an Intel Core i9-11900 CPU with a base clock speed of 2.00 GHz, 32GB of RAM and an NVIDIA GeForce RTX 4090 GPU. The operating system is Windows 11 and the development environment includes Anaconda distribution version 23.10.0 with Python 3.9.19.

\subsubsection{Training datasets}
\label{sssc:ta}
The training set used for the experiments in this paper is MS1MV3 \citep{Deng2021}, which has been widely used in recent years in face recognition solutions \citep{Zhang2022}. MS1MV3 is based on the MS-Celeb-1M dataset \citep{Guo2016}, which contains approximately 10M images and 100k identities. Since the original version has many noisy images, the refined version is preprocessed by RetinaFace \citep{Deng2019retinaface}. After removing the noisy images, MS1MV3 includes 93k identities and 5.1M face images. The numbers of images and identities in the different face datasets used for training and evaluation are shown in Tab. \ref{tab:data}.

\begin{table}[!t]
    \centering
    \scriptsize 
    \caption{The numbers of identities and images in the training set, validation set and evaluation set. The last column represents the average number of face images per identity for each corresponding dataset.}
    \begin{tabular*}{\linewidth}{@{\extracolsep{\fill}}cccc}
        \toprule
        \textbf{Dataset} & \textbf{Identities} & \textbf{Images}  & \textbf{Images / Identitie} \\
        \midrule
        \midrule
        MS1MV3           & 93k                 & 5.1M             & 54  \\
        \midrule
        LFW              & 5749                & 13233            & 2   \\
        AgeDB-30         & 568                 & 16488            & 29  \\
        CALFW            & 5749                & 11652            & 2   \\
        CPLFW            & 5749                & 12174            & 2   \\
        CFP-FP           & 500                 & 7000             & 14  \\
        VGG2-FP          & 9131                & 3.31M            & 362 \\
        \midrule
        IJB-B            & 1845                & 76.8k            & 41  \\
        IJB-C            & 3531                & 148.8k           & 42  \\
        \bottomrule
    \end{tabular*}
    \label{tab:data}
\end{table}

\subsubsection{Evaluation benchmarks}
\label{sssc:eb}
To demonstrate the effectiveness of MSConv on face recognition and compare it with other approaches, we evaluate the trained models on eight different evaluation benchmarks as show in. These eight evaluation benchmarks are (1) Labeled Faces in the Wild (LFW) \cite{Huang2008}: A classic benchmark for face recognition with images from various environments, now near accuracy saturation, used to assess basic recognition performance. (2) AgeDB-30 \cite{Moschoglou2017}: Tests the model's ability to handle age variations with images from different age groups. (3) Cross-Age LFW (CALFW) \cite{Zheng2017}: Assesses recognition performance under age differences, focusing on capturing facial features despite aging. (4) Cross-Pose LFW (CPLFW) \cite{Zheng2018}: Measures robustness to pose variations, with significant differences in face angles across images. (5) Celebrities in Frontal-Profile in the Wild (CFP-FP) \cite{Sengupta2016}: Evaluates performance in matching frontal and profile faces, challenging due to significant angle differences. (6) Visual Geometry Group Face $2$ in Frontal-Profile (VGGFace2-FP) \cite{Cao2018}: A large-scale set from diverse settings, used to evaluate model performance in complex and varied scenarios. (7) IARPA Janus Benchmark-B (IJB-B) \cite{Whitelam2017}, and (8) IARPA Janus Benchmark-C (IJB-C) \cite{Maze2018}. IJB-B evaluates model performance under varying poses, lighting, and expressions. IJB-C is an extended version with a larger scale and higher difficulty, further testing the model's robustness in complex scenarios. Refer to Table \ref{tab:data} for the amount of information contained in these six datasets. For the first six evaluation benchmarks, we report the accuracies of the models trained with different loss functions to evaluate their effectiveness. For the latter two evaluation benchmarks, IJB-B and IJB-C, we calculated the True Accept Rate at False Accept Rate(TAR@FAR) metric, where FAR was set to $1e-4$ and $1e-5$, respectively. 

\subsection{Hyperparameter Study}
\label{ssc:bb}
In this section, we will explore the impact of different batch sizes, epochs, learning rates, and dilation sizes on the performance of MSConv.

\subsubsection{Batch Size}
\label{sssc:bs}
This experiment selected batch sizes of 32, 64, 128, and 256 to evaluate the impact of different batch sizes on the final precision and loss of the model. The experimental results are shown in Table \ref{tab:bs}. The experimental model remained stable under different batch sizes. As the batch size increased, the training speed improved significantly, and the loss gradually decreased. The accuracy on some validation sets, such as LFW and CFP-FP, also increased. While performance on the AgeDB, IJB-B, and IJB-C datasets remained relatively stable, increasing the batch size helped the model converge faster and improved its generalization ability on certain tasks. For this study, a batch size of 128 was selected. 

\begin{table}[!t]
    \centering
    \caption{Performance of MSConv-R50 comparison under different batch size. The train epoch is set to 5. The accuracy on the IJB-B and IJB-C datasets is obtained at FAR=1e-4.}
    \resizebox{\linewidth}{!}{
    \begin{tabular}{c|c|c|c|c|c|c|c}
        \hline
         \textbf{Batch Size} & \makecell{\textbf{Speed} \\ \textbf{(Samples/Second)}} & \textbf{Loss} & \textbf{\makecell{LFW\\(\%)}} & \textbf{\makecell{CFP-FP\\(\%)}}& \textbf{\makecell{AgeDB\\(\%)}}& \textbf{\makecell{IJB-B(\%) \\ @FAR=1e-4}}& \textbf{\makecell{IJB-C(\%) \\ @FAR=1e-4}}\\
         \hline \hline
         32  &  670 &  7.6517 &  99.63 &  95.21 &  96.72 &  92.70 & 94.53 \\
         64  &  905 &  6.3589 &  99.68 &  96.28 &  97.08 &  93.25 & 94.95 \\
         128 &  918 &  6.2193 &  99.70 &  96.77 &  96.68 &  93.24 & 94.91 \\
         256 &  925 &  6.1995 &  99.58 &  96.39 &  96.12 &  92.96 & 94.65 \\
         \hline
    \end{tabular}
    }
    \label{tab:bs}
\end{table}

\begin{table}[!t]
    \centering
    \caption{Performance of MSConv-R50 is compared with different learning rate. The train epoch is set to 20.}
    \resizebox{\linewidth}{!}{
    \begin{tabular}{c|c|c|c|c|c}
        \hline
         \textbf{Learning Rates} & \textbf{LFW(\%)} & \textbf{CFP-FP(\%)} & \textbf{AgeDB(\%)} & \textbf{\makecell{IJB-B(\%) \\ @FAR=1e-4}} & \textbf{\makecell{IJB-C(\%) \\ @FAR=1e-4}}\\
         \hline \hline
         0.1  &  99.72 &  97.04&  97.45 &  93.89 & 95.51 \\
         0.02 &  99.83 &  97.94&  97.83 &  94.68 & 96.04 \\
         0.01 &  99.77 &  98.03&  97.68 &  94.46 & 95.80 \\
         \hline
    \end{tabular}
    }
    \label{tab:lr}
\end{table}

\subsubsection{Learning Rate}
\label{sssc:lr}
With the batch size fixed at 128, we compared the accuracy of three different learning rates, which were 0.1, 0.02, and 0.01. The experimental results are shown in Table \ref{tab:lr}. To optimize the training process, we adopted a cosine annealing learning rate adjustment method. This method gradually decreases the learning rate following a cosine function, decaying from the initial learning rate to a minimum value of 5e-6. Specifically, the learning rate changes follow a curve consistent with the cosine function in the interval [0, $\pi$/2]. The results show that when the learning rate is set to 0.02, the experimental model achieves higher accuracy on the validation sets LFW and AgeDB, as well as the test sets IJB-B and IJB-C, compared to when the learning rate is set to 0.1 or 0.001. Therefore, 0.02 is employed as the learning rate
in the paper.

\subsubsection{Epoch}
\label{sssc:epc}
Using a fixed batch size of 128 and a learning rate of 0.02, we compared the model's accuracy after 5, 20, and 25 epochs. By examining the accuracy at these epochs, we aimed to understand the trade-off between training time and model performance. This approach allowed us to determine whether additional training epochs significantly enhance the model's accuracy or if comparable results can be achieved with shorter training periods. Table \ref{tab:ep} presents the model performance differences at these three training epochs. The results indicate that increasing the number of training epochs improves model performance, with the experimental model demonstrating better accuracy after 25 epochs. However, as the number of training epochs increases, so does the required training time. From the results, we observe that the performance difference between 20 and 25 epochs is not significant. Therefore, we chose 20 epochs to reduce training time.
\begin{table}[!t]
    \centering
    \caption{Performance of MSConv-R50 is compared with different batch size.}
    \resizebox{\linewidth}{!}{
    \begin{tabular}{c|c|c|c|c|c|c}
        \hline
         \textbf{Epoch} & \textbf{Loss} & \textbf{LFW(\%)} & \textbf{CFP-FP(\%)}&  \textbf{AgeDB(\%)}&  \textbf{\makecell{IJB-B(\%) \\ @FAR=1e-4}}& \textbf{\makecell{IJB-C(\%) \\ @FAR=1e-4}} \\
        \hline 
        \hline
          5  & 6.2193 & 99.70  & 96.77 & 96.68 & 93.24 & 94.91\\
          20 & 2.2107 & 99.83  & 97.94 & 97.83 & 94.68 & 96.04\\
          25 & 2.0156 & 99.80  & 97.76 & 97.93 & 94.94 & 96.21\\
        \hline
    \end{tabular}
    }
    \label{tab:ep}
\end{table}

\subsubsection{The dilation D}
\label{sssc:d}
The dilation D is a crucial elements to control the receptive field (RF) size. To study its effects, we start from the two-branch case and fix the setting 3 $\times$ 3 filter with dilation D = 1 in the first kernel branch of MSConv. We use three different kernels, called K3 (standard 3 $\times$ 3 convolutional kernel), K5 (3 $\times$ 3 convolution with dilation 2 to approximate 5 $\times$ 5 kernel size) and K7 (3 $\times$ 3 with dilation 3 to approximate 7 $\times$ 7 kernel size). Although the kernel sizes in the table are not consistent, the number of parameters is the same. Considering that while larger convolution kernels have a larger receptive field, they also increase the number of parameters and computational complexity, we have used dilated convolutions to replace standard convolutions.

\begin{table}[!t]
    \centering
    \caption{Results of MSConv-R50 with different combinations of multiple kernels, where epoch and learning rate are 20 and 0.02, respectively.}
    \resizebox{\linewidth}{!}{
    \begin{tabular}{c|c|c|c|c|c|c|c|c}
         \hline
         \textbf{Index} &
         \textbf{\makecell{K3 \\ (D=1)}} & \textbf{\makecell{K5 \\ (D=2)}} & \textbf{\makecell{K7 \\ (D=3)}} & \textbf{LFW(\%)} &\textbf{CFP-FP(\%)} & \textbf{AgeDB-30(\%)} & \textbf{\makecell{IJB-B(\%) \\ @FAR=1e-4}} & \textbf{\makecell{IJB-C(\%) \\ @FAR=1e-4}}\\
         \hline \hline
         \multirow{2}{*}{1} & \checkmark & & & \multirow{2}{*}{99.8} & \multirow{2}{*}{97.44} & \multirow{2}{*}{97.88} & \multirow{2}{*}{94.29} & \multirow{2}{*}{95.86}\\ \cline{2-4}
         \multirow{2}{*}{} & \checkmark & & & & & & \\ \hline
         \multirow{2}{*}{2} & \checkmark & & & \multirow{2}{*}{\textbf{99.83}} & \multirow{2}{*}{\textbf{97.94}} & \multirow{2}{*}{\textbf{97.83}} & \multirow{2}{*}{\textbf{94.68}} & \multirow{2}{*}{\textbf{96.04}}\\ \cline{2-4}
         \multirow{2}{*}{} & & \checkmark & & & & & & \\ \hline
         \multirow{2}{*}{3} & & \checkmark & & \multirow{2}{*}{99.82} & \multirow{2}{*}{97.69} & \multirow{2}{*}{97.80} &\multirow{2}{*}{94.46} & \multirow{2}{*}{95.84}\\ \cline{2-4}
         \multirow{2}{*}{} & \checkmark & & & & & & & \\ \hline
         \multirow{2}{*}{4} & & \checkmark & & \multirow{2}{*}{99.77} & \multirow{2}{*}{97.60} & \multirow{2}{*}{97.63} & \multirow{2}{*}{94.46} & \multirow{2}{*}{94.46}\\ \cline{2-4}
         \multirow{2}{*}{} & & & \checkmark & & & & & \\ 
         \hline
    \end{tabular}
    }
    \label{tab:dd}
\end{table}

Table \ref{tab:dd} shows that the optimal combination is K3+K5 (Index 2), which performs significantly better than the combination using two kernels of the same size, K3+K3 (Index 1), and is the best-performing combination among the four different combinations shown in the table. It is proved beneficial to use different kernel sizes, and we attribute the reason to the aggregation of multi-scale information. Index 2 (K3 + K5) is the optimal combination because it can better balance the extraction of local and more extensive contextual information. This multi-scale feature extraction method provides local detail feature extraction through K3 while utilizing the K5 convolution kernel to expand the receptive field and capture broader contextual information. This combination performs excellently in both simple and complex scenarios, especially on the complex datasets IJB-B and IJB-C, making it suitable for a wide range of face recognition tasks. Index 1 (K3 + K3) shows average performance and is somewhat insufficient in handling complex scenarios, particularly due to its lack of effective ability to capture broader contextual information. Index 3 (K5 + K3) combination is similar to Index 2, but its performance is inferior to Index 2. This indicates that, on one hand, subtracting the features extracted by the larger convolution kernel (K5) from those extracted by the smaller convolution kernel (K3) can better demonstrate the advantages of SO. On the other hand, the features extracted by the larger convolution kernel are used as effective supplementary information and added to the features fused with the attention mechanism. If the features extracted by a smaller convolution kernel are used as supplementary information, it may introduce redundancy. Index 4 (K5 + K7) performs the worst, worse than the Index 3 combination, demonstrating that solely relying on larger convolution kernels to capture more global features cannot compensate for the lack of local detail feature extraction.

\subsection{Ablation Studies}
\label{ssc:dd}
In this section, we conduct ablation studies to inspect the relative effectiveness of different components in the proposed MSConv. We choose ResNet50 as the backbone by replacing MO or SO to conduct the following ablation experiments on MS1MV3.

\label{sssc:mo}
Table \ref{tab:ms} shows the performance changes after removing or replacing MO and SO in MSConv. When both MO and SO are present (MO $\&$ SO), the model achieves the best performance on most datasets. This indicates that the combination of MO and SO works well together for multi-scale feature extraction, effectively capturing both local details and broader contextual information. Removing SO alone has a greater negative impact on the model compared to removing MO alone. This demonstrates that our proposed method of using SO to extract difference features is highly effective. Removing both MO and SO results in worse performance than removing MO alone, but it performs slightly better than removing SO alone. This suggests that SO plays an irreplaceable role and also helps MO perform better. Replacing MO with SUM in MSConv does not achieve better results, failing to surpass the benefits of using MO. Overall, this experiment demonstrates that the combination of MO and SO is effective in improving model performance.

\subsection{Accuracy of Different Approaches}
\label{ssc:aa}
In this section, we used ResNet50 as the backbone to compare the performance of MSConv with six other different models on MS1MV3. To ensure fairness, all models were trained from scratch for 20 epochs using the SGD optimizer, with an initial learning rate of 0.02 and a batch size of 128. 

\begin{table}[!t]
    \centering
    \caption{Experimental results were obtained using different combination methods of MO and SO on the MS1MV3 dataset. Note that MSConv* indicates using element-wise addition (SUM) to replace MO while keeping SO unchanged.} 
    \resizebox{\linewidth}{!}{
    \begin{tabular}{l|c|c|c|c|c|c|c}
        \hline
         \textbf{Description} & \textbf{LFW(\%)} & \textbf{CFP-FP(\%)} & \textbf{AgeDB(\%)} & \textbf{\makecell{IJB-B(\%) \\ @FAR=1e-4}} &\textbf{\makecell{IJB-B(\%) \\ @FAR=1e-5}} & \textbf{\makecell{IJB-C(\%) \\ @FAR=1e-4}} & \textbf{\makecell{IJB-C(\%) \\ @FAR=1e-5}}\\
         \hline \hline
         MSConv -- MO         & 99.67            & 96.66                & 96.45              & 93.04          & 87.51          & \textbf{94.99} & \textbf{92.16}\\
         MSConv -- SO         & 98.97            & 89.19                & 91.22              & 90.48          & 79.95          & 92.50          & 86.93\\
         MSConv -- MO -- SO   & 99.50            & 92.30                & 94.75              & 90.71          & 81.79          & 92.79          & 87.55\\
         MSConv*              & 99.67            & 96.63                & \textbf{96.72}     & 92.99          & 87.52          & 94.73          & 87.35\\
         MSConv               & \textbf{99.70}   & \textbf{96.77}       & 96.68              & \textbf{93.24} & \textbf{88.35} & 94.91          & 92.11\\
         \hline
    \end{tabular} 
    } 
    \label{tab:ms}
\end{table}

\textbf{Results on LFW, CALFW, CPLFW, AgeDB-30, CFP-FP, and VGG2-FP.} 
To compare with recent state-of-the-art methods, we trained our model and six other models on the MS1MV3 dataset and evaluated them on various benchmarks. As reported in Table \ref{tab:all}, the proposed MSConv achieves state-of-the-art performance on the LFW, CFP-FP, AgeDB-30, and CALFW face verification datasets. MSConv attains the highest accuracy on LFW (99.83\%), demonstrating its strong baseline discriminative capability in unconstrained environments (e.g., varying illumination, partial occlusion, and natural poses). Its leading performance on CFP-FP (97.94\%) validates its adaptability to extreme pose variations (e.g., frontal vs. profile faces). The superior results on AgeDB-30 (97.83\%) indicate that MSConv effectively models long-term facial changes (e.g., wrinkle accumulation and bone structure variations), confirming its robustness to age differences. For CALFW, which combines age and pose challenges, MSConv’s leading accuracy (96.08\%) proves its ability to handle multi-factor interference synergistically. As analyzed earlier, MSConv is closely related to SKNet. Although SKNet slightly outperforms MSConv by 0.17\% on CPLFW, MSConv surpasses SKNet on the remaining five validation sets with improvements of 0.03\%, 0.05\%, 0.05\%, 0.03\%, and 0.18\%, respectively, demonstrating the effectiveness of our enhancements over SKNet. On VGG2-FP, GhostNet’s accuracy is only 0.25\% higher than MSConv (ranked second), while MSConv outperforms GhostNet in other metrics, confirming its overall superiority.
We further evaluated MSConv’s stability using different loss functions (CML, Cos, Arc). While CML yields the best overall performance, the results highlight the impact of loss function selection on accuracy. A suitable loss function can better guide MSConv to achieve balanced performance across multiple tasks.

\begin{table}[!t]
    \centering
    \caption{The accuracies of different models on the validation set. Respectively, CML, Cos and Arc express Combined Margin Loss, CosFace and ArcFace. Note that, FasterNet-S does not use ResNet50 as the backbone but instead employs the structure closest to ResNet50 from the original paper, with a depth of [1, 2, 13, 2].}
    \resizebox{\textwidth}{!}{
    \begin{tabular}{l|c|c|c|c|c|c}
    \hline
    \textbf{Method} & \textbf{LFW(\%)} & \textbf{CFP-FP(\%)} & \textbf{AgeDB-30(\%)} & \textbf{CALFW(\%)} & \textbf{CPLFW(\%)} & \textbf{VGG2-FP(\%)}\\ 
    \hline \hline
    SCConv \citep{Li2023}           & 99.43           & 85.59          & 95.20          & 94.90          & 86.50          & 87.22\\
    FasterNet--S \citep{Chen2023}   & 99.70           & 93.70          & 96.85          & 95.72          & 90.05          &92.72\\
    GhostNet \citep{Han2020}        & 99.78           & 97.13          & 97.68          & 96.02          & 91.63          &\textbf{95.14}\\
    GhostNetV2 \citep{Tang2022}     & 99.73           & 96.24          & 97.30          & 95.75          & 91.68          &94.2\\  
    SKNet \citep{Li2019}            & 99.80           & 97.89          & 97.78          & 96.05          & \textbf{93.02} & 94.72\\ 
    StarNet \citep{Ma2024}          & 98.85           & 82.84          & 91.67          & 93.00          & 82.82          &85.4\\
    \hline \hline 
    MSConv--CML                    & \textbf{99.83}  & \textbf{97.94} & \textbf{97.83} & \textbf{96.08} & 92.85          &94.9\\
    MSConv--Cos                    & 99.77           & 97.64          & 97.95          & 96.18          & 92.60          &94.7\\
    MSConv--Arc                    & 99.80           & 97.67          & 97.98          & 96.12          & 92.77          &94.84\\
    \hline
    \end{tabular}
    }
    
    \label{tab:all}
\end{table}

\textbf{Results on IJB-B, IJB-C.} 
The performance of all seven models on the test set is shown in Table \ref{tab:bc}, with MSConv ranking first on both the IJB-B and IJB-C benchmarks. On the IJB-B benchmark, MSConv achieves a TAR of 94.68 at FAR = 1e-4 and 90.34 at FAR = 1e-5, outperforming other state-of-the-art models in both cases. On the IJB-C benchmark, MSConv achieves a TAR of 96.04 at FAR = 1e-4 and 93.91 at FAR = 1e-5, consistently outperforming other models. The SKNet is the most related work to ours. However, under the strict false accept rate condition (FAR = 1e-5) on the IJB-B benchmark, MSConv achieves a 0.55\% improvement in TAR (90.34\% vs. 89.79\%) at the cost of a 17.8\% increase in parameters (1.39M vs. 1.18M) and a 19.2\% increase in FLOPs (31.6M vs. 26.5M). Similarly, on the IJB-C benchmark under FAR=1e-5, MSConv outperforms SKNet by 0.05\% in TAR (93.91\% vs. 93.86\%). Without significantly increasing computational resource consumption, MSConv enhances recognition robustness in high-security scenarios and low-quality environments, demonstrating its superior adaptability. Although the choice of loss function affects validation set performance, MSConv consistently outperforms comparative models on test sets regardless of the loss function used, demonstrating its task-agnostic adaptability without dependence on specific loss functions.

\begin{table}[!t]
    \centering
    \caption{Compare the performance of different models on the IJB-B and IJB-C datasets, including TAR (True Accept Rate) at FAR (False Accept Rate) thresholds of 1e-4 and 1e-5, and \#Param (number of trainable parameters) and FLOPs (Floating Point Operations). The first six comparison methods use CML as the loss function.}
    \resizebox{\textwidth}{!}{
    \begin{tabular}{lcccccc}
        \toprule 
        \multirow{2}{*}{\textbf{Method}} & \multicolumn{2}{c}{\textbf{IJB-B(\%)}} & \multicolumn{2}{c}{\textbf{IJB-C(\%)}} & \multirow{2}{*}{\textbf{\#Param(G)}} & \multirow{2}{*}{\textbf{FLOPs(M)}} \\
        \cmidrule(lr){2-3} \cmidrule(lr){4-5}
        & \textbf{\makecell{TAR \\ @ FAR = 1e-4}} & \textbf{\makecell{TAR \\ @ FAR = 1e-5}} & \textbf{\makecell{TAR \\ @ FAR = 1e-4}} & \textbf{\makecell{TAR \\ @ FAR = 1e-5}} \\
        \toprule 
        SCConv         & 87.48          & 77.41          & 90.80          & 84.08          & 6.89          & 46.8 \\
        FasterNet--S   & 92.57          & 86.94          & 94.42          & 91.47          & 1.10          & 30.6 \\
        GhostNet       & 94.14          & 90.19          & 95.72          & 93.43          & 1.69          & 22.0 \\
        GhostNetV2     & 93.42          & 88.00          & 95.10          & 92.38          & 0.62          & 14.3 \\
        SKNet          & 94.57          & 89.79          & 96.01          & 93.86          & 1.18          & 26.5 \\
        StarNet        & 73.11          & 51.90          & 74.26          & 58.34          & \textbf{0.59} & \textbf{13.6} \\
        \toprule
        MSConv--CML    & \textbf{94.68} & \textbf{90.34} & \textbf{96.04} & \textbf{93.91} & 1.39          & 31.6 \\
        MSConv--Cos    & 94.88          & 90.79          & 96.17          & 94.07          & 1.39          & 31.6 \\
        MSConv--Arc    & 94.69          & 89.76          & 96.07          & 93.97          & 1.39          & 31.6 \\
        \bottomrule
        \end{tabular}
        }
    \label{tab:bc}
\end{table}

\subsection{Visualization.}
\begin{figure}[!t]
	\centering
	\includegraphics[width=0.5\columnwidth]{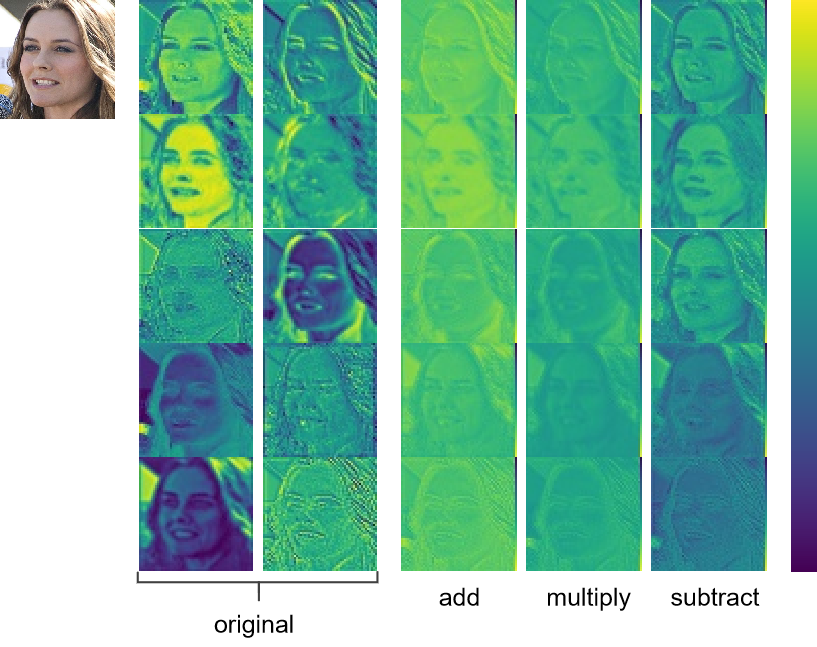}
	\caption{original: Visualization of features after the same input passes through 3×3 and 5×5 convolutions, respectively. "add", "multiply", and "subtract" represent element-wise addition, multiplication, and subtraction, respectively.}
	\label{fg:figure6}
\end{figure}

In order to investigate the roles of the MO and SO in the proposed MSConv module, we compare the feature representations of two original feature groups after applying element-wise addition, subtraction, and multiplication in Fig. \ref{fg:figure6}. The two feature groups are selected from the outputs of the dual-scale convolutional layer in the first module of stage 1 in the MSConv-embedded ResNet50 model. We visualize the top five feature channels out of the total 64 channels. It can be observed that, compared to feature addition, the features extracted by MO exhibit greater saliency, while SO is more adept at capturing fine-grained differences in the feature maps.

\begin{figure}[!t]
	\centering
	\includegraphics[width=0.55\columnwidth]{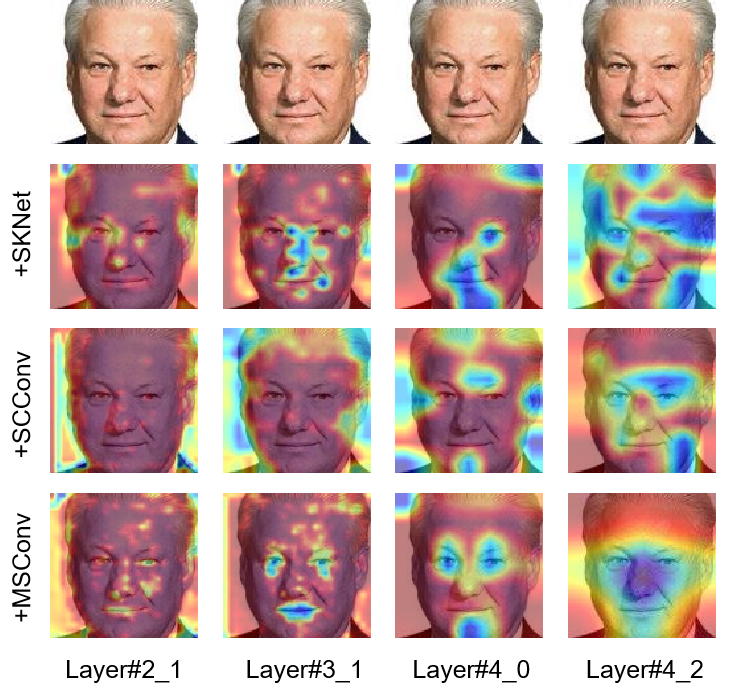}
	\caption{Visualization of class activation mapping at different layers using ResNet50 as backbone networks.}
	\label{fg:gram}
\end{figure}

To provide a more intuitive demonstration of the effectiveness of the MSConv, we sample 1 image from VGG2-FP validation dataset and utilize GradCAM \citep{selvaraju2017grad} to visualize the class activation mapping. For comparative analysis, we also generate heat maps for SKNet and SCConv embedded models. The visualizations are presented in Fig. \ref{fg:gram} and the models all trained on MS1MV3. The result displays attention heat maps from four different layers, illustrating MSConv’s impact in refning attention and positioning accurately. As a result, the activation map becomes more precise, capturing critical and accurate locations for semantic representations.

\begin{figure}[!t]
	\centering
	\includegraphics[width=0.5\columnwidth]{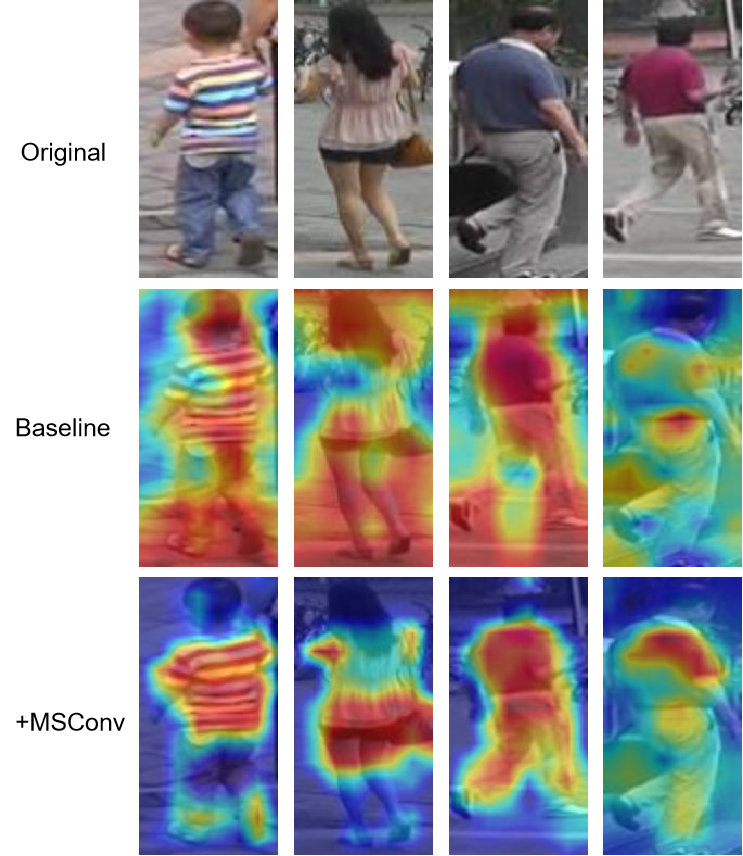}
	\caption{Comparison of attention regions for pedestrian pose estimation between the baseline (ResNet) and MSConv-embedded ResNet.}
	\label{fg:figure7}
\end{figure}

Additionally, we applied the MSConv structure to the person re-identification task to test its generalization ability, and visualized the results as shown in Fig. \ref{fg:figure7}. After integrating the MSConv module into the base model ResNet, the experimental results show that MSConv significantly enhances the model's attention to different person postures, allowing it to focus more effectively on key regions and avoid interference from background information.

\section{Conclusion}
\label{sc:conc}
In this paper, we address the critical issue in face recognition where existing feature fusion methods overly rely on salient features while neglecting differential features. We propose a novel convolutional module, MSConv, designed to balance these two types of features. Specifically, MSConv integrates multi-scale hybrid convolution. MO dynamically weights features without additional parameters to amplify salient patterns, while SO suppreass redundant noise and enhance differential features. Experimental results on benchmark datasets demonstrate that MSConv significantly outperforms models focusing solely on salient features. Current limitations include the unexplored generalizability of MSConv to other computer vision tasks, such as low-resolution or age-invariant face recognition. Future research can further expand the generalization capability of MSConv in various tasks such as object detection, medical image analysis, and remote sensing image processing, while designing more task-specific model variants to enhance its adaptability and performance.

\section*{Acknowledgements}
\label{sc:ack}
This work was supported in part by the Guangdong Basic and Applied Basic Research Foundation under Grant 2022A515110020, the National Natural Science Foundation of China under Grant 62271232, and the National Natural Science Foundation of China under Grant 62106085.

\bibliographystyle{elsarticle-harv} 
\bibliography{reference}

\end{document}